\documentclass[runningheads]{llncs}

\usepackage{epsfig}
\usepackage{graphicx}
\usepackage{amsmath}
\usepackage{amssymb}
\usepackage{wrapfig}

\def\eg{e.g.,\ }
\def\ie{i.e.,\ }

\newcommand{\cartoondataset}{CartoonSet}
\newcommand{\xgan}{\textsc{xgan}}
\newcommand{\ldann}{\mathcal{L}_{dann}}
\newcommand{\cdann}{c_{dann}}
\newcommand{\lconstst}{{\mathcal L}_{sem, 1 \rightarrow 2}}
\newcommand{\lconstts}{{\mathcal L}_{sem, 2 \rightarrow 1}}
\newcommand{\lganst}{{\mathcal L}_{gan, 1 \rightarrow 2}}
\newcommand{\lgants}{{\mathcal L}_{gan, 2 \rightarrow 1}}
\newcommand{\lteach}{{\mathcal L}_{teach}}
\newcommand{\dom}{\mathcal D}
\newcommand{\enc}{e}
\newcommand{\dec}{d}
\newcommand{\genst}{g_{1 \rightarrow 2}}
\newcommand{\gents}{g_{2 \rightarrow 1}}

\begin{document}
\author{Am\'{e}lie Royer\inst{1}\orcidID{0000-0002-8407-0705} \and
Konstantinos Bousmalis\inst{2, 6}\and
Stephan Gouws\inst{2} \and
Fred Bertsch\inst{3} \and
Inbar Mosseri\inst{4} \and
Forrester Cole\inst{4} \and
Kevin Murphy\inst{5}}
\authorrunning{A. Royer et al.}

\institute{IST Austria, 3400 Klosterneuburg, Austria\\
Work done while at Google Brain London, UK\\
\email{aroyer@ist.ac.at}
\and
Google Brain, London, UK\\
\email{\{konstantinos, sgouws\}@google.com}
\and
Google Brain, Mountain View, USA\\
\and
Google Research, Cambridge, USA\\
\and
Google Research, Mountain View, USA\\
\and 
Currently at Deepmind, London, UK
}

\title{{XGAN}: Unsupervised {I}mage-to-{I}mage {T}ranslation for {M}any-to-{M}any {M}appings}
\titlerunning{{XGAN}: Unsupervised {I}mage-to-{I}mage {T}ranslation }
\maketitle

 \begin{abstract}
 Image translation refers to the task of mapping images from a visual domain to another. Given two unpaired collections of images, we aim to learn a mapping between the corpus-level style of each collection, while preserving semantic content shared across the two domains.
 We introduce \xgan{}, a dual adversarial auto-encoder, which captures a shared representation of the common domain semantic content in an unsupervised way, while jointly learning the
 domain-to-domain image translations in both directions. 
 We exploit ideas from the domain adaptation literature and define a \textit{semantic consistency loss} which encourages the learned embedding to preserve semantics shared across domains.
 We report promising qualitative results for the task of face-to-cartoon translation. The cartoon dataset we collected for this purpose, "CartoonSet", is also publicly available as a new benchmark for semantic style transfer at \url{https://google.github.io/cartoonset/index.html}.

 \end{abstract}

\section{Introduction}

Image-to-image translation -- learning to map images from one domain to another -- covers several classical computer vision tasks such as style transfer (rendering an image in the style of a given input~\cite{Gatys2015c}), colorization (mapping grayscale images to color images~\cite{zhang2016colorful}), super-resolution (increasing the resolution of an input image~\cite{ledig2016photo}), or semantic segmentation (inferring pixel-wise semantic labeling of a scene~\cite{fcn}).
Learning such mappings requires an underlying understanding of the shared information between the two domains. 
In many cases, supervision encapsulates this knowledge in the form of labels or paired samples. This holds for instance for colorization, where ground-truth pairs are easily obtained by generating grayscale images from colored inputs.

\begin{figure}[!hbt]
\begin{center}
\includegraphics[height=3.4cm]{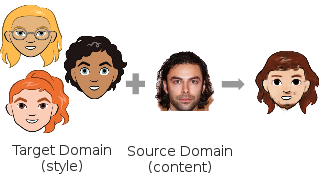}~~~~~~~\includegraphics[height=3.4cm]{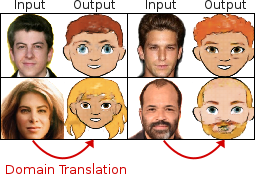}
  \end{center}
  \caption{\label{fig:semantic-ss} Semantic style transfer is the task of adapting an image to the visual appearance of another domain without altering its semantic content given only two unpaired image collections without pairs supervision (\textit{left}). 
  We define semantic content as characteristic attributes which are shared across domains, but do not necessarily appear the same at the pixel-level.
  For instance, cartoons and faces have a similar range of hair color but with very different appearances, \eg blonde hair is bright yellow in cartoons.
  The proposed \xgan{} applied on the face-to-cartoon task yields a shared representation that preserves important face semantics such as hair style or face shape (\textit{right}). }
\end{figure}

In this work, we consider the task of \textit{unsupervised semantic style transfer}: learning to map an image from one domain into the style of another domain without altering its semantic content (see Figure~\ref{fig:semantic-ss}). 
In particular, we experiment on the task of translating faces to cartoons.
Note that without loss of generality, a photo of a face can be mapped to many valid cartoons, and vice-versa. 
Semantic style transfer is therefore a \emph{many-to-many mapping} problem, for which obtaining labeled examples is ambiguous and costly. 
Furthermore in this unsupervised setting we do not have access to supervision on shared domain semantic content  (e.g., facial attributes such as hair color, eye color, etc.). Instead, we propose an encoder-decoder structure with a bottleneck embedding shared across the two domains to capture common semantics as a latent representation. 

The key issue is thus to learn an embedding that preserves semantic facial attributes (hair color, eye color, etc.) between the two domains with little supervision, and to incorporate it within a generative model to produce the actual domain translations. 
Although this paper specifically focuses on the face-to-cartoon setting, many other examples fall under this category: mapping landscape pictures to paintings (where the different scene objects and their composition describe the input semantics), transforming sketches to images, or even cross-domain tasks such as generating images from text. 
We only rely on two unlabeled training image collections or \textit{corpora}, one for each domain, with no known image pairings across domains.
Hence, we are faced with a double \textit{domain shift}, first in terms of global domain appearance, and second in terms of the content distribution of the two collections.

Recent work~\cite{discogan,cyclegan,dualgan,pixelda,cycada} report good performance using GAN-based models for unsupervised image-to-image translation when the two input domains share similar pixel-level structure (\eg horses and zebras) but fail for more significant domain shifts (\eg dogs and cats).
Perhaps the best known recent example is CycleGAN~\cite{cyclegan}. Given two image domains $\dom_1$ and $\dom_2$, the model is trained with a pixel-level \emph{cycle-consistency loss} which ensures that the mapping $\genst$ from $\dom_1$ to $\dom_2$ followed by its inverse, $\gents$, yields the identity function; \ie $\genst \circ \gents = id$.
We argue that such a pixel-level constraint is not sufficient in our setting, and that we rather need a constraint in \emph{feature space} to allow for more permissive transformations of the pixel input. 
To this end, we propose \xgan{} (``Cross-GAN''), a dual adversarial auto-encoder which learns a shared semantic representation of the two input domains in an unsupervised way, while jointly learning both domain-to-domain translations. 
More specifically, the domain translation $\genst$ consists of an encoder $e_1$ taking inputs in $\dom_1$, followed by a decoder $d_2$ with outputs in $\dom_2$ (and likewise for $\gents$) such that $e_1$ and $e_2$, as well as $d_1$ and $d_2$, are partially shared across domains.

The main novelty lies in how we constrain the shared embedding using techniques from the domain adaptation literature, as well as a novel \textit{semantic consistency loss}. The latter ensures that the domain-to-domain translations preserve the semantic representation, \ie that $e_1 \approx e_2 \circ \genst$ and $e_2 \approx e_1 \circ \gents$. 
Therefore, it acts as a form of self-supervision which alleviates the need for paired examples and preserves semantic feature-level information rather than pixel-level content.
In the following section, we review relevant recent work before discussing the \xgan{} model in more detail in Section \ref{sec:proposed-model}. In Section~\ref{sec:dataset}, we introduce \textsc{\cartoondataset{}}, our dataset of cartoon faces for research on semantic style transfer. Finally, in Section~\ref{sec:experiments} we report experimental results of \xgan{} on the face-to-cartoon task.

\section{Related work}
\label{sec:related}

Recent literature suggests two main directions for tackling the semantic style transfer task: traditional style transfer and pixel-level domain adaptation.
The first approach is inadequate as it only transfers texture information from a single style image, and therefore does not capture the style of an entire corpus.
The latter category also fails in practice as it explicitly enforces pixel-level similarity which does not allow for significant structural change of the input.
Instead, we draw inspiration from the domain adaptation and feature-level image-to-image translation literature.

\paragraph{Style Transfer.} Neural style transfer refers to the task of transferring the texture of a \textit{specific} style image while preserving the pixel-level structure of an input content image~\cite{Gatys2015c,Johnson2016}. 
Recently,~\cite{LiW16,Liao:2017:VAT:3072959.3073683} proposed to instead use a dense local patch-based matching approach in the feature space, as opposed to global feature matching, allowing for convincing  transformations between visually dissimilar domains.
Still, these models only perform image-specific transfer rather than learning a global \textit{corpus-level} style and  do not provide a meaningful shared domain representation.
Furthermore, the generated images are usually very close to the original input in terms of pixel structure (\eg edges) which is not suitable for drastic transformations such as face-to-cartoon.

\paragraph{Domain adaptation.} \xgan{} relies on learning a shared feature representation of both domains in an unsupervised setting to capture semantic rather than pixel information. For this purpose, we make use of the domain-adversarial training scheme~\cite{dann}.
Moreover, recent domain adaptation work~\cite{dsn,shrivastava2017learning,pixelda} can be framed as semantic style transfer as they tackle the problem of mapping synthetic images, easy to generate, to natural images, which are more difficult to obtain. 
The generated samples are then used to train a model later applied to natural images.
Contrary to our work however, they only consider pixel-level transformations.

\paragraph{Unsupervised Image-to-Image translation.} Recent work~\cite{discogan,cyclegan,dualgan,cycada} tackle the unsupervised pixel-level image-to-image translation task by learning both cross-domain mappings jointly, each as a separate generative adversarial network, via a cycle-consistency loss which ensures that applying each mapping followed by its reverse yields the identity function.
This intuitive form of self-supervision leads to good results for pixel-level transformations, but often fails to capture significant structural changes \cite{cyclegan}.
In comparison, our proposed semantic consistency loss acts at the feature-level, allowing for more flexible transformations.

Orthogonal to this line of work is UNIT~\cite{unit,munit,egunit}. This model consists of a coupled VAEGAN architecture~\cite{vaegan,cogan} with a shared embedding bottleneck, trained with pixel-level cycle-consistency. Similar to \xgan{}, it learns a joint \textit{feature-level} representation of the two domains, however UNIT assumes that sharing high-level layers in the architecture is a sufficient constraint, while \xgan{}'s objective explicitly introduces the semantic consistency component.

Finally, the \textit{Domain Transfer Network} (DTN)~\cite{dtn,wolf2017unsupervised} is closest to our work in terms of objective and applications. The DTN architecture is a single auto-encoder trained to map images from a source to a target domain with self-supervised semantic consistency feedback.
It was also successfully applied to the problem of feature-level image-to-image translation, in particular to the face-to-cartoon problem.
Contrary to \xgan{} however, the DTN encoder is pretrained and fixed, and is assumed to produce meaningful embeddings for both the face and the cartoon domains.
This assumption is very restrictive, as off-the-shelf models pretrained on natural images do not usually generalize well to other domains.
In fact, we show in Section~\ref{sec:experiments} that a fixed encoder does not generalize well in the presence of a large domain shift between the two domains.

\section{Proposed model: XGAN}
\label{sec:proposed-model}

\begin{figure*}[!htb]
\begin{center}
  \begin{minipage}[t]{0.29\textwidth}
  \centering \includegraphics[height=3.1cm]{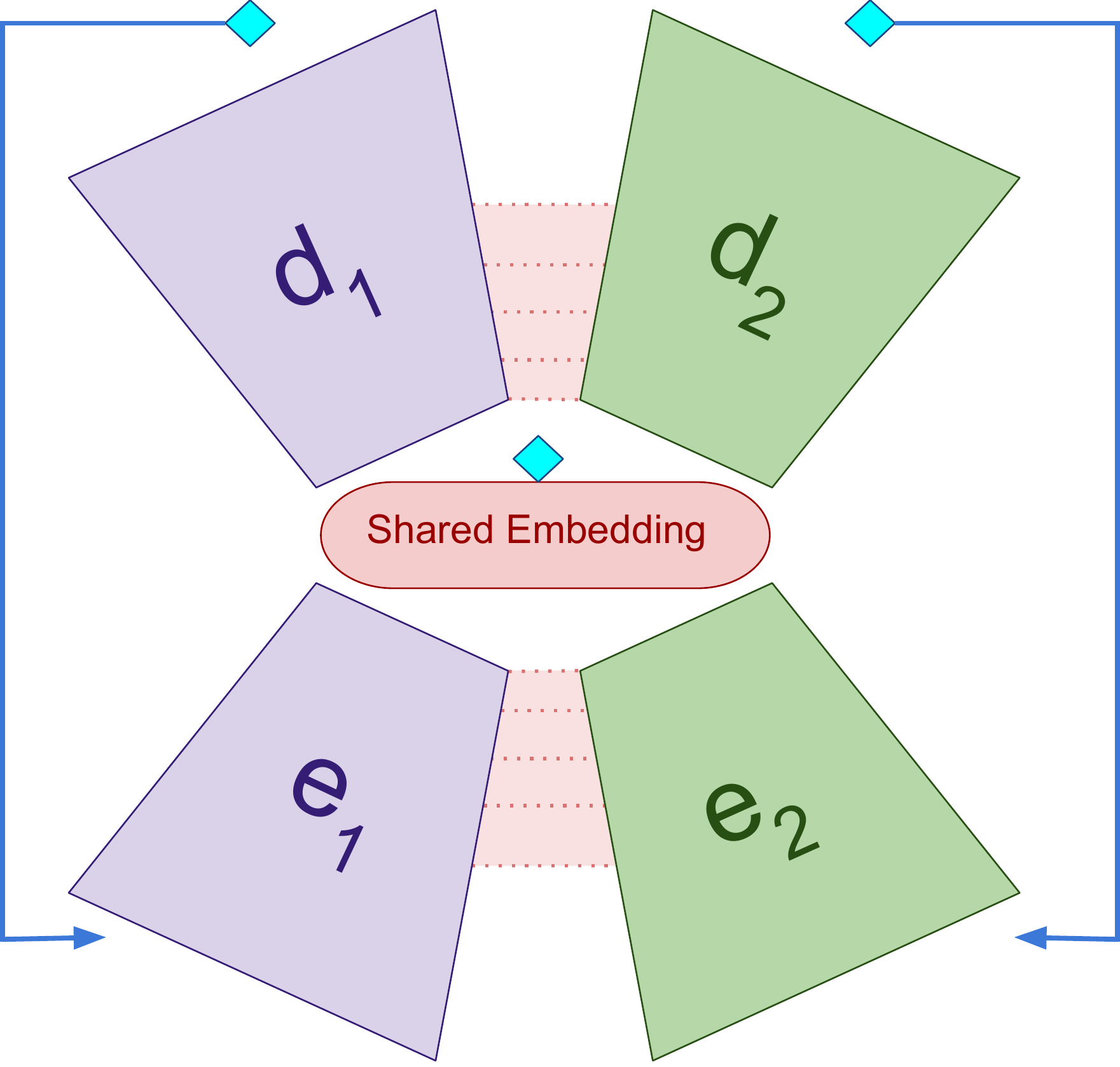}
  
  \centering \footnotesize (A) High-level view of the \xgan{} dual auto-encoder architecture
  \end{minipage}~~
  \begin{minipage}[t]{0.225\textwidth}
  \centering \includegraphics[height=3.1cm]{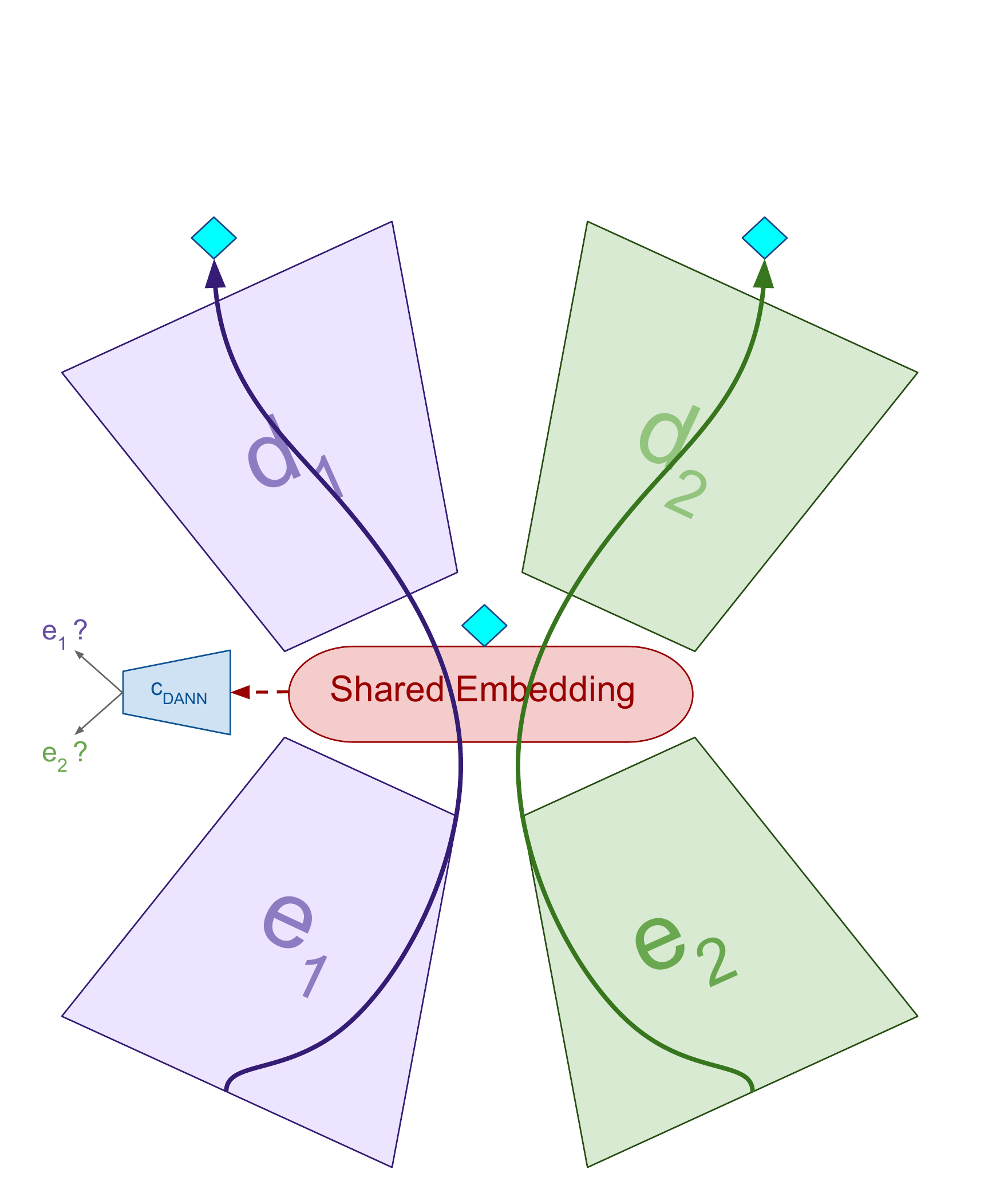}
  
  \centering \footnotesize (B1) Domain-adversarial auto-encoder
  \end{minipage}~
  \begin{minipage}[t]{0.22\textwidth}
  \centering \includegraphics[height=3.1cm]{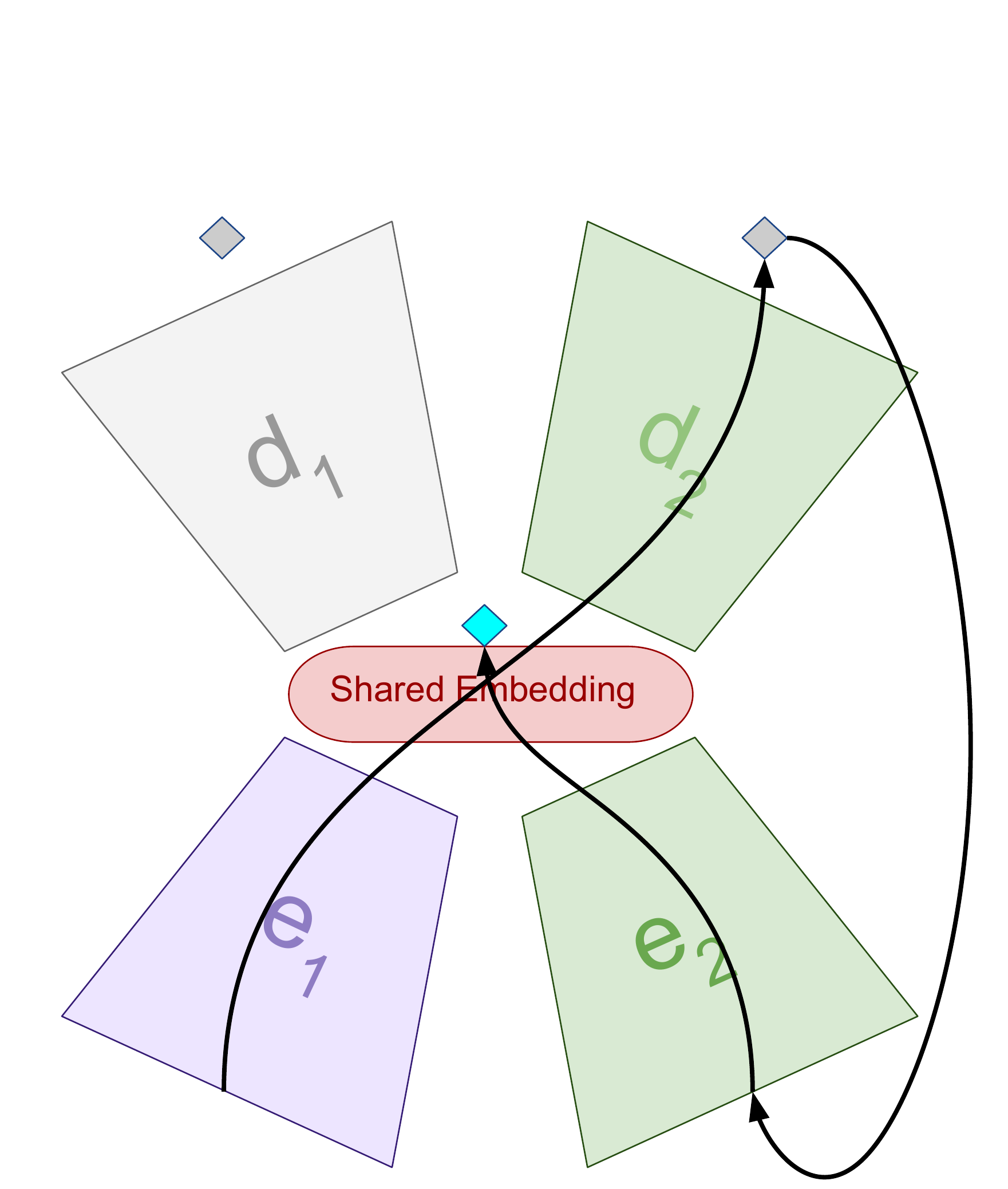}
  
  \centering \footnotesize (B2) Semantic consistency feedback loop
  \end{minipage}~~
  \begin{minipage}[t]{0.23\textwidth}
  \centering \includegraphics[height=3.1cm]{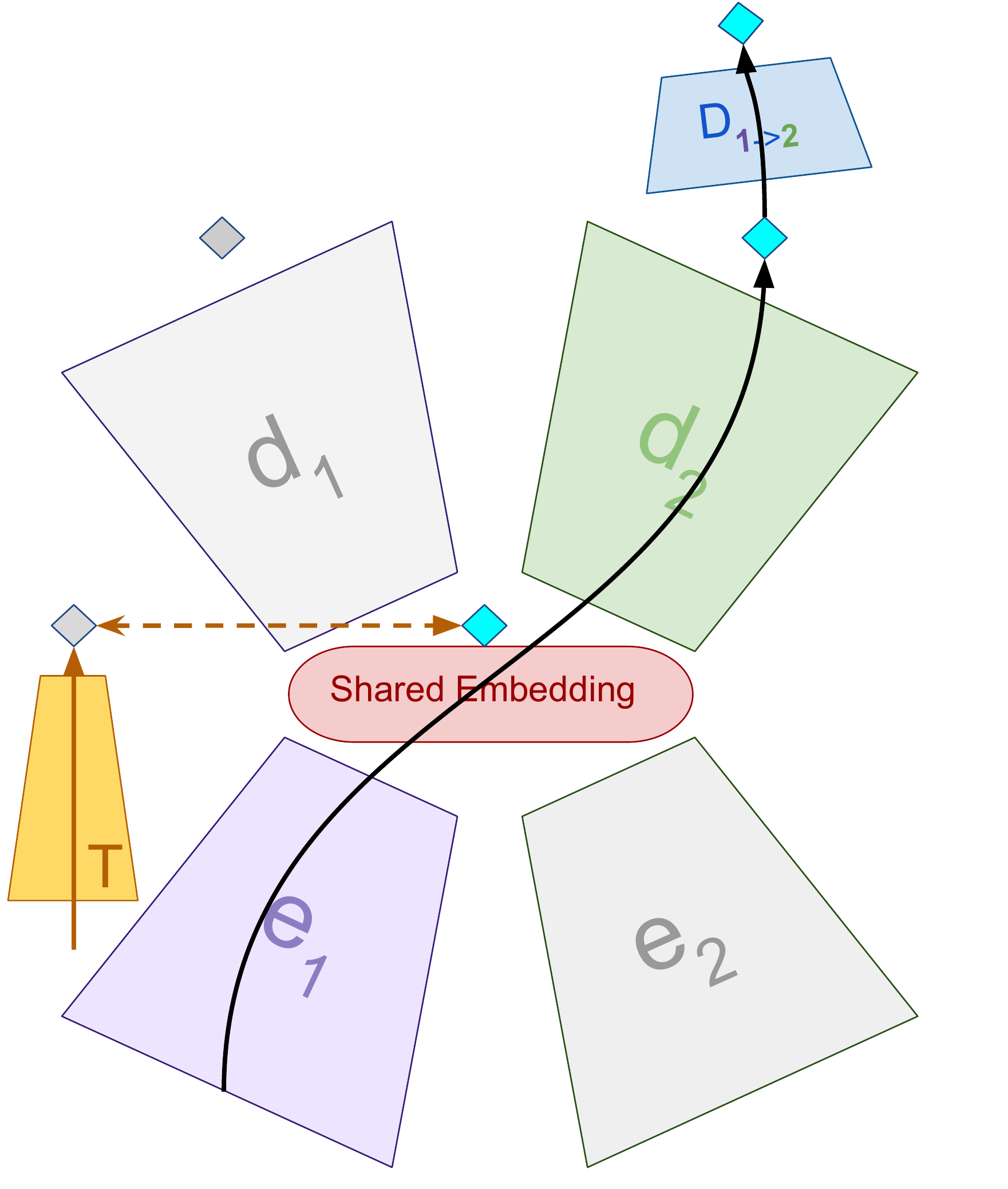}
  
  \centering \footnotesize (B3) GAN and Teacher loss modules
  \end{minipage}
  \end{center}
  \caption{\label{fig:xgan}The \xgan{} (A) objective encourages the model to learn a meaningful joint embedding (B1) ($\mathcal{L}_{rec}$ and $\mathcal{L}_{dann}$), which should be preserved through domain translation (B2) ($\mathcal{L}_{sem}$), while producing output samples of good quality (B3) ($\mathcal{L}_{gan}$ and $\lteach$)}
\end{figure*}

Let $\dom_1$ and $\dom_2$ be two domains that differ in terms of \textit{visual appearance} but share common \textit{semantic content}.
It is often easier to think of domain semantics as a high-level notion, e.g., semantic attributes, however we do not require such annotations in practice, but instead consider learning a feature-level representation that automatically captures these shared semantics.
Our goal is thus to learn in an unsupervised fashion, \ie without paired examples, a joint domain-invariant embedding: semantically similar inputs across domains will be embedded nearby in the learned feature space.
Architecture-wise, \xgan{} is a dual auto-encoder  on domains $\dom_1$ and $\dom_2$  Figure \ref{fig:xgan}(A). We denote by $\enc_1$ the encoder and by $\dec_1$  the decoder for domain $\dom_1$; likewise $\enc_2$ and $\dec_2$ for $\dom_2$. 
For simplicity, we also denote by $\genst = \dec_2 \circ \enc_1$ the transformation from $\dom_1$ to $\dom_2$; likewise $\gents$ for $\dom_2$ to $\dom_1$.

The training objective can be decomposed into five main components: 
the \textit{reconstruction} loss, $\mathcal{L}_{rec}$, encourages the learned embedding to encode meaningful knowledge for each domain; 
the \textit{domain-adversarial} loss, $\ldann$, pushes embeddings from $\dom_1$ and $\dom_2$ to lie in the same subspace, bridging the domain gap at the semantic level;
the \textit{semantic consistency} loss, $\mathcal{L}_{sem}$, ensures that input semantics are preserved after domain translation; 
$\mathcal{L}_{gan}$ is a simple generative adversarial (GAN) objective,  encouraging the model to generate more realistic samples, and finally, $\lteach$ is an optional teacher loss that distills prior knowledge from a fixed pretrained teacher embedding, when available.
The total loss function is defined as a weighted sum over these five loss terms:
\begin{align*}
\label{eq:xgan}
    {\mathcal L}_{\xgan} =  \mathcal{L}_{rec} + \omega_{d} \mathcal{L}_{dann} + \omega_{s} \mathcal{L}_{sem} + \omega_{g} \mathcal{L}_{gan} + \omega_{t} \lteach, 
\end{align*}
where the $\mathbf{\omega}$ hyper-parameters control the contributions from each of the individual objectives. 
An overview of the model is given in Figure \ref{fig:xgan}, and we discuss each objective in more detail in the rest of this section.

\paragraph{Reconstruction loss, $\mathcal{L}_{rec}$.} $\mathcal{L}_{rec}$  encourages the model to encode enough information on each domain for to perfectly reconstruct the input. More specifically $\mathcal{L}_{rec} = \mathcal{L}_{rec, 1} + \mathcal{L}_{rec, 2}$ is the sum of reconstruction losses for each domain. 
    \begin{align}
        \mathcal{L}_{rec, 1} = \mathbb{E}_{\mathbf{x} \sim p_{\dom_1}} \left( \| \mathbf{x} - \dec_1(\enc_1(\mathbf{x}))\|_2 \right),\ \mbox{\small likewise for domain } \mathcal{D}_{2} 
    \end{align}

\paragraph{Domain-adversarial loss, $\ldann$.} $\ldann$ is the domain-adversarial loss between $\dom_1$ and $\dom_2$, as introduced in \cite{dann}. It encourages the embeddings learned by $\enc_1$ and $\enc_2$ to lie in the same subspace. In particular, it guarantees the soundness of the cross-domain transformations $\genst$ and $\gents$.
More formally, this is achieved by training a binary classifier, $\cdann$, on top of the embedding layer to categorize encoded images from \emph{both} domains as coming from either $\dom_1$ or $\dom_2$ (see Figure \ref{fig:xgan} (B1)).
    $\cdann$ is trained to maximize its classification accuracy while the encoders $\enc_1$ and $\enc_2$ simultaneously strive to minimize it, \ie to confuse the domain-adversarial classifier.
    Denoting model parameters by $\theta$ and a classification loss function by $\ell$ (\eg cross-entropy), we optimize
    \begin{align}
       &\min_{\theta_{e_1}, \theta_{e_2}} \max_{\theta_{dann}} \mathcal{L}_{dann} \mbox{,   where } \\
       &\mathcal{L}_{dann} = \mathbb{E}_{p_{\dom_1}} \ell(1, c_{dann}(e_1(\mathbf{x}))) + \mathbb{E}_{p_{\dom_2}} \ell\left(2, c_{dann}(e_2(\mathbf{x}))\right) \nonumber
    \end{align}

\paragraph{Semantic consistency loss, $\mathcal{L}_{sem}$.}
Our key contribution is a semantic consistency feedback loop that acts as self-supervision for the cross-domain translations $\genst$ and $\gents$. 
Intuitively, we want the semantics of input $\mathbf{x} \in \dom_1$  to be preserved when translated to the other domain, $\genst(\mathbf{x}) \in \dom_2$, and similarly for the reverse mapping.
However this consistency property is hard to assess at the pixel-level as we do not have paired data and pixel-level metrics are sub-optimal for image comparison.
Instead, we introduce a feature-level semantic consistency loss, which encourages the network to preserve the learned embedding during domain translation. Formally, $\mathcal{L}_{sem} = \lconstst + \lconstts$, where: 
\begin{align}
    &\lconstst = \mathbb{E}_{\mathbf{x} \sim p_{\dom_1}} \| \enc_1 (\mathbf{x}) - \enc_2(\genst(\mathbf{x})) \|, 
    \mbox{likewise for }\lconstts.\\
    &\| \cdot \| \mbox{ denotes a distance between vectors. } \nonumber
\end{align}

\paragraph{GAN objective, $\mathcal{L}_{gan}$.}
We find that generating realistic image transformations has a crucial positive effect for learning a joint meaningful and semantically consistent embedding as the produced samples are fed back through the encoders when computing the semantic consistency loss: making the transformed distribution $p(\gents(\dom_2))$ as close as possible to the original domain $p(\dom_1)$ ensures that the encoder $\enc_1$ does not have to cope with an additional domain shift.

Thus, to improve sample quality, we add a generative adversarial loss~\cite{gan} $\mathcal{L}_{gan} = \lganst + \lgants$, where $\lganst$ is a state-of-the-art GAN objective~\cite{gan} where the generator $\genst$ is paired against the discriminator $D_{1 \rightarrow 2}$ (and likewise for $\gents$ and $D_{2 \rightarrow 1}$). 
    In this scheme, a discriminator $D_{1 \rightarrow 2}$ strives to distinguish generated samples from real ones in $\dom_2$, while the generator $\genst$ aims to produce samples that confuse the discriminator. The formal objective is
    \begin{align}
    &\min_{\theta_{g_{1 \rightarrow 2}}}  \max_{\theta_{D_{1 \rightarrow 2}}} \lganst  
\\
    &{\lganst = \mathbb{E}_{\mathbf{x} \sim p_{\dom_2}} \left( \log(D_{1 \rightarrow 2}(\mathbf{x})) \right) + \mathbb{E}_{\mathbf{x} \sim p_{\dom_1}} \left( \log(1 - D_{1 \rightarrow 2}(\genst(\mathbf{x}))) \right)} \nonumber
    \end{align}
Likewise $\lgants$ is defined for the transformation from $\dom_2$ to $\dom_1$.

Note that the combination of the $\mathcal{L}_{gan}$ and $\mathcal{L}_{sem}$ objectives should subsume the role of the domain-adversarial loss $\ldann$ in theory. However, $\ldann$ plays an important role at the beginning of training to bring embeddings across domains closer, as the generated samples are typically poor and not yet representative of the actual input domains $\dom_1$ and $\dom_2$.

\paragraph{Teacher loss, $\lteach$.} We introduce an optional component to  incorporate prior knowledge in the model when available, \eg in a semi-supervised setting.
$\lteach$  encourages the learned embeddings to lie in a region of the subspace defined by the output representation of a  pretrained teacher network, $T$.
    In other words, we distills feature-level knowledge from $T$ and constrains the embeddings to a more meaningful sub-region, relative to the task on which $T$ was trained; This can be seen as a form of regularization of the learned embedding.
    Moreover, $\lteach$ is asymmetric by definition. It should not be used for both domains simultaneously as each term would potentially push the learned embedding in two different directions. Formally, $\lteach$ (applied to domain $\dom_1$) is defined as:

    \begin{align}
        \lteach = &\  \mathbb{E}_{\mathbf{x} \sim p_{\dom_1}} \| T(\mathbf{x}) - \enc_1(\mathbf{x}) \|,\\ &\mbox{ where $\| \cdot \|$ is a distance between vectors.} \nonumber
    \end{align}

\subsection{Architecture and Training procedure}
\label{sec:arc}

We use a simple mirrored convolutional architecture for the auto-encoder. It consists of 5 convolutional blocks for each encoder, the two last ones being shared across domains, and likewise for the decoders (5 deconvolutional blocks with the two first ones shared). 
This encourages the model to learn shared representations at different levels of the architecture rather than only in the middle layer.
A more detailed description is given in Table \ref{tab:arch}.
For the teacher network, we use the highest convolutional layer of FaceNet~\cite{Schroff_2015_CVPR}, a state-of-the-art face recognition model trained on natural images.

The \xgan{} training objective is to minimize (Eq. \ref{eq:xgan}). In particular, the two adversarial losses ($\mathcal{L}_{gan}$ and $\mathcal{L}_{dann}$) lead to min-max optimization problems requiring careful optimization.
For the GAN loss $\mathcal{L}_{gan}$, we use a standard adversarial training scheme~\cite{gan}.
Furthermore, for simplicity we only use one discriminator in practice, namely $D_{1 \rightarrow 2}$ which corresponds to the face-to-cartoon path, our target application.
We first update the parameters of the generators $\genst$ and $\gents$ in one step. We then keep these fixed and update the parameters for the discriminator $D_{1 \rightarrow 2}$. We iterate this alternating process throughout training.
The adversarial training scheme for $\ldann$ can be implemented in practice by connecting the classifier $\cdann$ and the embedding layer \emph{via} a gradient reversal layer~\cite{dann}: the feed-forward pass is unaffected, however the gradient is backpropagated to the encoders with a sign-inversion representing the min-max alternation.
We perform this update simultaneously when computing the generator parameters.
Finally, we train the model with \textsc{Adam} optimizer~\cite{adam} and an initial learning rate of $1\mbox{e-}4$.

\begin{table}[tb]
    \centering
    
\makebox[0pt][c]{\parbox{\textwidth}{%
    \begin{minipage}[t]{0.3\textwidth}
    \begin{center}
    \begin{tabular}{|l|c|}
    \hline
        Layer & Size \\
        \hline
        \hline
        Inputs &  64x64x3\\
        \hline
        conv1 & 32x32x32 \\
        \hline
        conv2 & 16x16x64 \\
        \hline
        (\textbf{//}) conv3 & 8x8x128 \\
        \hline
        (\textbf{//}) conv4 & 4x4x256 \\
        \hline
        (\textbf{//}) FC1 & 1x1x1024 \\
        \hline
        (\textbf{//}) FC2 & 1x1x1024 \\
        \hline
    \end{tabular}
    
    (a) Encoder
    \end{center}
    \end{minipage}~~~~
    \begin{minipage}[t]{0.31\textwidth}
    \begin{center}
    \begin{tabular}{|c|c|}
    \hline
        Layer & Size \\
        \hline
        \hline
        Inputs & 1x1x1024\\
        \hline
        (//) deconv1 & 4x4x512 \\
        \hline
        (//) deconv2 & 8x8x256 \\
        \hline
        deconv3 & 16x16x128 \\
        \hline
        deconv4 & 32x32x64 \\
        \hline
        deconv5 & 64x64x3 \\
        \hline
        \multicolumn{1}{c}{}\\
        \multicolumn{1}{c}{}\\
    \end{tabular}
    
    (b) Decoder
    \end{center}
    \end{minipage}~~~~
    \begin{minipage}[t]{0.31\textwidth}
    \begin{center}
    \begin{tabular}{|c|c|}
    \hline
        Layer & Size \\
        \hline
        \hline
        Inputs &  64x64x3\\
        \hline
        conv1 & 32x32x16 \\
        \hline
        conv2 & 16x16x32 \\
        \hline
        conv3 & 8x8x32 \\
        \hline
        conv4 & 4x4x32 \\
        \hline
        FC1 & 1x1x1 \\
        \hline
        \multicolumn{1}{c}{}\\
        \multicolumn{1}{c}{}\\
    \end{tabular}
    
    (c) Discriminator
    \end{center}
    \end{minipage}
    
    }}
    \caption{\label{tab:arch}Overview of the XGAN architecture used in practice. The encoder and decoder have the same architecture for both domains, and (\textbf{//}) indicates that the layer is shared across domain.}
\end{table}

\section[cartoon]{The \cartoondataset{} Dataset}
\label{sec:dataset}
Although previous work has tackled the task of transforming frontal faces to a specific cartoon style, there is currently no such dataset publicly available. 
For this purpose, we introduce a new dataset, \cartoondataset{}\footnote{CartoonSet, \url{https://github.com/google/cartoonset}}, which we release publicly to further aid research on this topic.

Each cartoon face is composed of 16 components including 12 facial attributes (\eg facial hair, eye shape, etc) and 4 color attributes (such as skin or hair color) which are chosen from a discrete set of RGB values. %
The number of options per attribute category ranges from 3 to 111, for the largest category, hairstyle. 
Each of these components and their variation were drawn by the same artist, resulting in approximately 250 cartoon components artworks and $10^8$ possible combinations.
The artwork components are divided into a fixed set of layers that define a Z-ordering for rendering.
For instance, face shape is defined on a layer below eyes and glasses, so that the artworks are rendered in the correct order.
For instance, hair style needs to be defined on two layers, one behind the face and one in front. There are 8 total layers: hair back, face, hair front, eyes, eyebrows, mouth, facial hair, and glasses.
The mapping from attribute to artwork is also defined by the artist such that any random selection of attributes produces a visually appealing cartoon without any misaligned artwork; which sometimes involves handling interaction between attributes, e.g. the appearance of "short beard" will changed depending of the face shape.
For example, the proper way to display a "short beard" changes for different face shapes, which required the artist to create a "short beard" artwork for each face shape. 
We create the \cartoondataset{} dataset from arbitrary cartoon faces by randomly sampling a value for each attribute. 
We then filter out unusual hair colors (pink, green etc) or unrealistic attribute combinations, which results in a final dataset of approximately $9,000$ cartoons.
In particular, the filtering step guarantees that the dataset only contains realistic cartoons, while being completely unrelated to the source dataset.
\begin{figure}[tbh]
    \centering \includegraphics[trim={0 11.35cm 0 0}, clip, width=.39\textwidth]{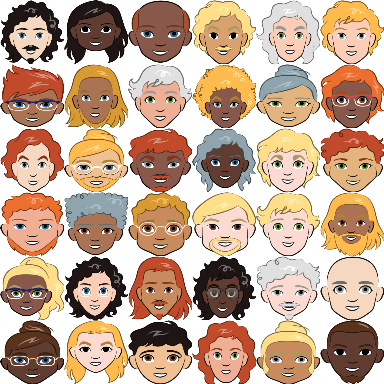}\includegraphics[trim={0 0 0 11.35cm}, clip, width=.39\textwidth]{Images/trg_samples.png}\includegraphics[trim={6.8cm 2.24cm 0 9.05cm}, clip, width=.2\textwidth]{Images/trg_samples.png}
  \caption{\label{fig:dataset} Random samples from our cartoon dataset, \cartoondataset{}. 
  }
\end{figure}

\begin{figure}[tbh]
    \centering \includegraphics[trim={0 11.35cm 0 0}, clip, width=.39\textwidth]{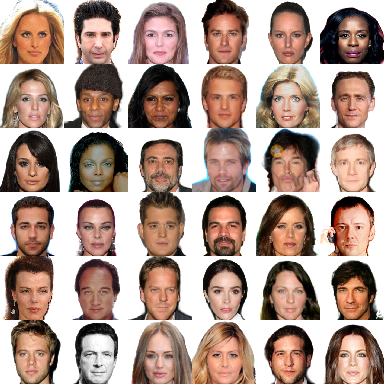}\includegraphics[trim={0 0 0 11.35cm}, clip, width=.39\textwidth]{Images/src_samples.png}\includegraphics[trim={6.8cm 2.24cm 0 9.05cm}, clip, width=.2\textwidth]{Images/src_samples.png}
  \caption{\label{fig:vgg_dataset} Random centered aligned samples from VGG-Face. We preprocess them with automatic portrait matting to avoid dealing with background noise.
  }
\end{figure}

\section{Experiments}
\label{sec:experiments}

We experimentally evaluate our \xgan{} model on \textit{semantic style transfer}; more specifically, on the task of converting images of frontal faces (source domain) to images of cartoon avatars (target domain) given an unpaired collection of such samples in each domain. 
Our source domain is composed of real-world frontal-face images from the VGG-Face dataset~\cite{Parkhi15}. In particular, we use an image collection consisting of 18,054 uncropped celebrity frontal face pictures. As a preprocessing step, we align the faces based on eyes and mouth location and remove the background.
The target domain is the CartoonSet dataset introduced in the previous section.
%
Finally, we randomly select and take out 20\% of the images from each dataset for testing purposes, and use the remaining 80\% for training. For our experiments we also resize all images to $64\times64$.
As shown in Figures \ref{fig:dataset} and \ref{fig:vgg_dataset}, the two domains vary significantly in appearance. 
In particular, cartoon faces are rather simplistic compared to real faces, and do not display as much variety (\eg noses or eyebrows only have a few shape options).
Furthermore, we observe a major \textit{ content distribution shift} between the two domains due to the way we collected the data: for instance, certain hair color shades (\eg bright red, gray) are over-represented in the cartoon domain compared to real faces. Similarly, the cartoon dataset contains many samples with eyeglasses while the source dataset only has a few.

\begin{figure*}[!htb]
\centering \includegraphics[width=0.98\textwidth]{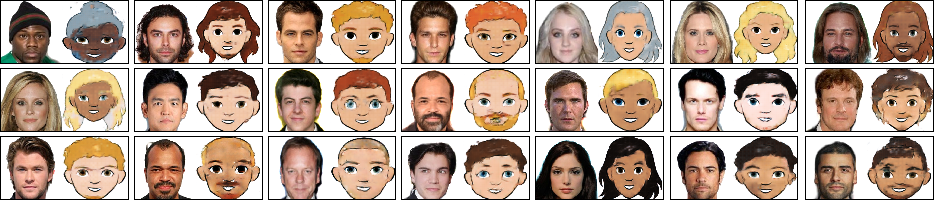}
\caption{\label{fig:ultraxgan} Selected samples generated by \xgan{} on the VGG-Face (left) to \cartoondataset{} (right) task. The figure reads row-wise: for each face-cartoon pair, the target image (cartoon) on the right was generated from the source image (face) on the left. }
\end{figure*}

\begin{figure}[!tbh]
\begin{center}
  \begin{minipage}[c]{0.95\textwidth}
    \centering \includegraphics[width=\textwidth]{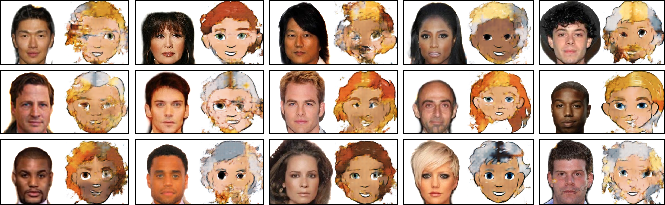}
    \centering \small (a) \textit{Baseline:} DTN
  \end{minipage}\quad
  
  \begin{minipage}[c]{0.95\textwidth}
    \centering \includegraphics[width=\textwidth]{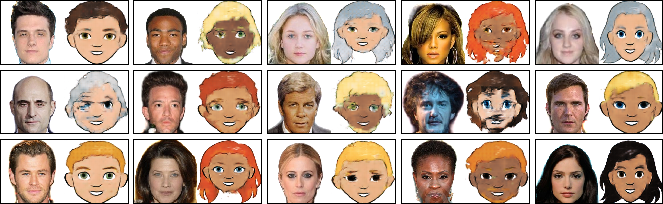}
    \centering \small (b) \textit{Proposed:}  XGAN
  \end{minipage}
  \end{center}
  \caption{A qualitative comparison between DTN and \xgan{}. In both cases we present random test samples for the face-to-cartoon transformation. The tables are organized row-wise where each face input is mapped to the cartoon face immediately on its right. 
  }
  \label{fig:dtn_vs_xgan}
\end{figure}

\paragraph{Comparison to the DTN baseline.} 
\label{sec:dtn}
Our first evaluation is a qualitative comparison between the Domain Transfer Network (DTN)~\cite{dtn} and \xgan{} on the semantic style transfer problem outlined above.
To the best of our knowledge, DTN is the current state of the art for semantic style transfer given unpaired image corpora from two domains with significant visual shift.
In particular, DTN was also applied to the task of transferring face pictures to cartoons (bitmojis) in the original paper\footnote{The original DTN code and dataset is not publicly available, hence we instead report results from our implementation applied to the VGG-Face to \cartoondataset{} setting.}. 
Figure \ref{fig:dtn_vs_xgan} shows the results of both DTN and \xgan{} applied to random VGG-Face samples from the test set to produce their cartoon counterpart. 
Evaluation metrics for style transfer are still an active research topic with no good unbiased solution yet. Hence we choose optimal hyperparameters by manually evaluating the quality of resulting samples, focusing on accurate transfer of semantic attributes, similarity of the resulting sample to the target domain, and crispness of samples.

\begin{figure}[tbh]
\centering \includegraphics[width=0.98\textwidth]{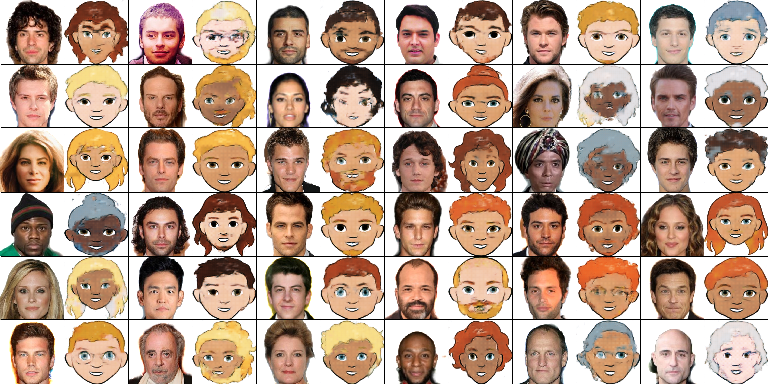}

\centering (a) Source to target mapping (face-to-cartoon)

\centering \includegraphics[width=0.98\textwidth]{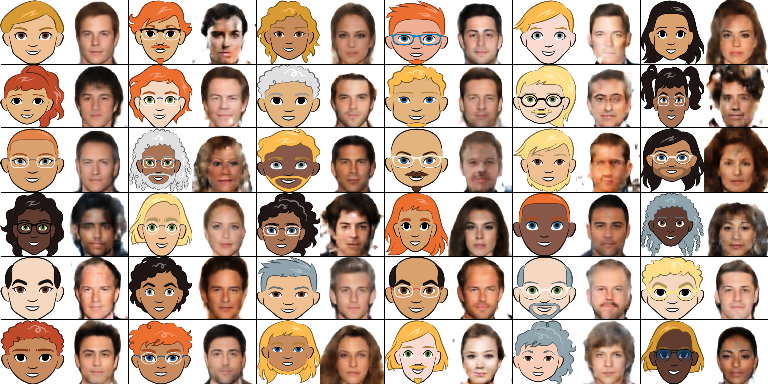}

\centering (b) Target to source mapping (cartoon-to-face)
\caption{\label{fig:qual}Random samples obtained by applying \xgan{} on faces and cartoons from the testing set for both cross-domain mappings}
\end{figure}

It is clear from Figure \ref{fig:dtn_vs_xgan} that DTN fails to capture the transformation function that semantically stylizes frontal faces to cartoons from our target domain. In contrast, XGAN is able to produce sensible cartoons both in terms of the style domain -- the resulting cartoons look crisp and respect the specific \cartoondataset{} style -- and in terms of semantic similarity to the input samples from VGG-Face. 
There are some failure cases such as hair or skin color mismatch, which emerge from the weakly supervised nature of the task and the significant content shift between the two domains (\eg red hair is over-represented in the target cartoon dataset).
In Figure \ref{fig:ultraxgan} we report selected \xgan{} samples that we think best illustrate its semantic consistency abilities, showing that the model learns a meaningful shared representation that preserves common face semantics.
Additional random samples are also reported in Figure \ref{fig:qual}.

We believe the failure of DTN is primarily due to its assumption of a fixed joint encoder for both domains.
Although the decoder learns to reconstruct inputs from the target domain almost perfectly, the semantics are not well preserved across domains and the decoder yields samples of poor quality for the domain transfer.
In fact, FaceNet was originally trained on real faces inputs, hence there is no guarantee it can produce a meaningful representation for \cartoondataset{} samples.
In contrast to our dataset, the target bitmoji domain in ~\cite{dtn} is visually closer to real faces, as bitmojis are more realistic and customizable than the cartoon style domain we use here. This might explain the original work performance even with a fixed encoder.
Our experiments suggest that using a fixed encoder is  too restrictive and does not adapt well to new scenarios.
We also train a DTN with a finetuned encoder which yields samples of better quality than the original DTN. However, this setup is very sensitive to hyperparameters choice during training and prone to mode collapse (see Section \ref{sec:dtnfine}).

\begin{figure}[tbh]
 \centering \includegraphics[width=0.95\textwidth]{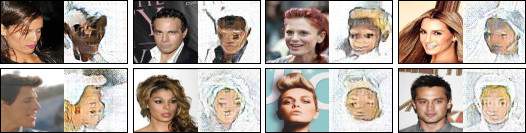}
 \caption{\label{cyclegan} The default CycleGAN model is not suitable for transformation between domains with very dissimilar appearances as it enforces pixel-level structural similarities}
 \end{figure}

 \paragraph{Comparison to CycleGAN.} As we have mentioned in the related work section, CycleGAN~\cite{cyclegan}, DiscoGAN~\cite{discogan} and DualGAN~\cite{dualgan} form another family of closely related work for image-to-image translation problems.
 However, differently from DTN and the proposed XGAN, these models only consider a pixel-level cycle consistency loss and do not use a shared domain embedding. Consequently, they fail to capture high-level shared semantics between significantly different domains.
 To explore this problem, we experiment with CycleGAN\footnote{CycleGAN-tensorflow, \url{https://github.com/xhujoy/CycleGAN-tensorflow}} on the face-to-cartoon task.
 We train a CycleGAN with a pix2pix~\cite{pix2pix} generator as in the original paper, which is close to the generator we use in XGAN in terms of architecture choices and size (depth and width of the network).
 As shown in Figure \ref{cyclegan}, this approach yields poor results, which is explained by the explicit pixel-level cycle consistency loss and the fact that the pix2pix architecture contains backwards connections (U-net) between the encoder and the decoder; both these features enhance pixel structure similarities which is not desirable for this task. 

\paragraph{Ablation study.}
\label{sec:abla}
We conduct a number of insightful ablation experiments on \xgan{}. 
We first consider training only with the reconstruction loss  $\mathcal{L}_{rec}$ and domain-adversarial loss $\mathcal{L}_{dann}$.
In fact these form the core domain adaptation component in \xgan{} and, as we will show, are already able to capture basic semantic knowledge across domains in practice.
Secondly we experiment with the semantic consistency loss and teacher loss. We show  that both have complementary constraining effects on the embedding space which contributes to improving the sample consistency.

\begin{figure}[!tbh]
\begin{center}
  \begin{minipage}[c]{0.95\textwidth}
    \centering \includegraphics[width=\textwidth]{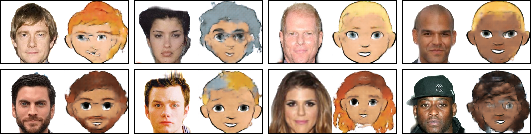}

    \centering \small (a) source to target
  \end{minipage}
  
  \begin{minipage}[c]{0.95\textwidth}
    \centering \includegraphics[width=\textwidth]{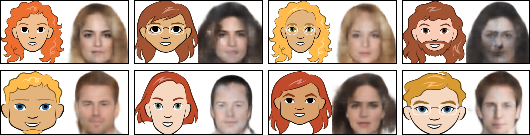}

    \centering \small (b) target to source
  \end{minipage}
  \end{center}
  \caption{\label{fig:dann} Test results for \xgan{} with the reconstruction ($\mathcal{L}_{rec}$) and domain-adversarial ($\ldann$) losses active only in the training objective $\mathcal L_{\xgan{}}$  }
\end{figure}

We first experiment on \xgan{} with only the reconstruction and domain-adversarial losses active.
These components prompt the model to (i) encode enough information for each decoder to correctly reconstruct images from the corresponding domain and (ii) to ensure that the embedding lies in a common subspace for both domains.
In practice in this setting, the model is robust to hyperparameter choice and does not require much tuning to converge to a good regime, \ie low reconstruction error and around 50\% accuracy for the domain-adversarial classifier.
As a result of (ii), applying each decoder to the output of the other domain's encoder yields reasonable cross-domain translations, albeit of low quality (see Figure \ref{fig:dann}).
Furthermore, we observe that some simple semantics such as skin tone or gender are overall well preserved by the learned embedding due to the shared auto-encoder structure.
For comparison, failure modes occur in extreme cases, \eg when the model capacity is too
small, in which case transferred samples are of poor quality, or when the weight $\omega_{d}$ is too low. In the latter case, the source and target embeddings are easily distinguishable and the cross-domain translations do not look realistic. 
Secondly, we investigate the benefits of adding semantic consistency in \xgan{} via the following three components:  \textit{sharing high-level layers} in the auto-encoder leads the model to capture common semantics earlier in the architecture.
In general, high-level layers in convolutional neural networks are known to encode semantic information.
 We performed experiments with sharing only the middle layer in the dual auto-encoder. As expected, the resulting embedding does not capture relevant shared domain semantics. 
 Second, we use the \textit{semantic consistency loss} as self-supervision for the learned embedding, ensuring that it is preserved through the cross-domain transformations.
It also reinforces the action of the domain-adversarial loss as it constrains embeddings from the two input domains to lie close to each other.
 Finally, the optional \textit{teacher loss} leads the learned source embedding to lie near the teacher output (in our case, FaceNet's representation layer), which is meaningful for real faces. It acts in conjunction with the domain-adversarial loss and semantic consistency loss, whose role is to bring the source and target embedding distributions closer to each other.

\begin{figure}[!tbh]
\begin{minipage}[c]{0.98\textwidth}
  \begin{minipage}[c]{0.49\textwidth}
    \centering \includegraphics[width=\textwidth]{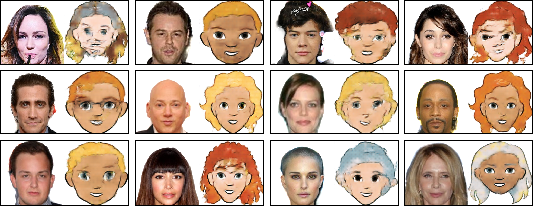}

    \centering \small (i) source to target
  \end{minipage}~
  \begin{minipage}[c]{0.49\textwidth}
    \centering \includegraphics[width=\textwidth]{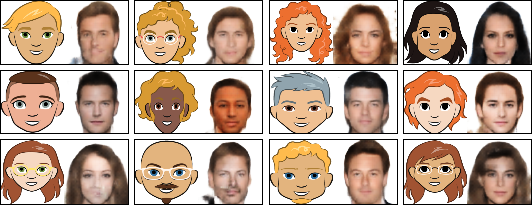}

    \centering \small (ii) target to source
  \end{minipage}
  
  \centering (a) Teacher loss inactive
  \end{minipage}
  
  \begin{minipage}[c]{0.98\textwidth}
  
  \begin{minipage}[c]{0.49\textwidth}
    \centering \includegraphics[width=\textwidth]{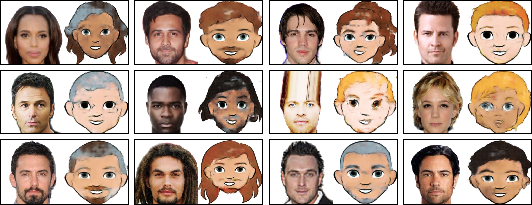}

    \centering \small (i) source to target
  \end{minipage}~
  \begin{minipage}[c]{0.49\textwidth}
    \centering \includegraphics[width=\textwidth]{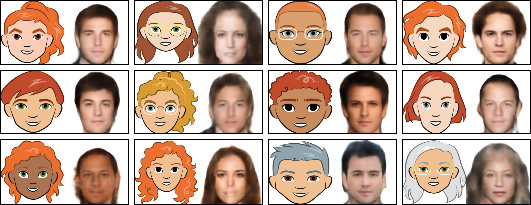}

    \centering \small (ii) target to source
  \end{minipage}
  
  \centering (b) Semantic consistency loss inactive
  \end{minipage}
  \caption{\label{fig:noconst} Results of ablating the teacher loss ($\lteach$) (top) and semantic consistency loss ($\mathcal{L}_{sem}$) (bottom) in the \xgan{} objective  $\mathcal L_{\xgan{}}$.}
\end{figure}

In Figure \ref{fig:noconst} we report random test samples for both domain translations when ablating the teacher loss and semantic consistency loss respectively.
While it is hard to draw conclusions from visual inspections, it seems that  the teacher network has a positive regularization effect on the learned embedding by guiding it to a more realistic region: training the model without the teacher loss (Figure \ref{fig:noconst} (a)) yields  more distorted samples, especially when the input is an outlier, \eg person wearing a hat, or cartoons with unusual hairstyle.
Conversely, when the semantic consistency is inactive (Figure \ref{fig:noconst} (b)), the generated samples overall display less variety. In particular, rare attributes (\eg unusual hairstyle) are not as well preserved as when the semantic consistency term is present.

\paragraph{Discussions and Limitations.}
Our initial aim was to tackle the \textit{semantic style transfer} problem in a fully unsupervised framework by combining techniques from domain adaptation and image-to-image translation:
We first observe that using a simple setup where a partially shared dual auto-encoder is trained with reconstruction  and domain-adversarial losses already suffices to produce an embedding that captures basic semantics rather well (for instance, skin tone). However, the generated samples are of poor quality and fine-grained attributes such as facial hair are not well captured.
These two problems are greatly diminished after adding the GAN loss and the proposed semantic consistency loss, respectively.
Failure cases still exist, especially on non-representative input samples (\eg a person wearing a hat) which are mapped to unrealistic cartoons.
Adding the teacher loss mitigates this problem by regularizing the learned embedding, however it requires additional supervision and makes the model dependent on the specific representation provided by the teacher network.
Future work will focus on evaluating \xgan{} on different domain transfer tasks.
In particular, though we introduced \xgan{} for semantic style transfer, we think the model goes beyond this scope  and  can be applied to classical domain adaptation problems, where quantitative evaluation becomes possible: while the pixel-level transformations are  not necessary for learning the shared embedding,  they are beneficial for learning a meaningful representation across visual domains, when combined with the self-supervised semantic consistency loop.

\section{Conclusions}
In this work, we introduced \xgan{}, a model for unsupervised domain translation applied to the task of semantically-consistent style transfer.
In particular, we argue that, similar to the domain adaptation task, learning image-to-image translation between two structurally different domains requires learning a high-level joint semantic representation while discarding local pixel-level dependencies. Additionally, we proposed a semantic consistency loss acting on both domain translations as a form of self-supervision.

We reported promising experimental results on the task of face-to-cartoon that outperform the current baseline.
We also showed that additional weak supervision, such as a pretrained feature representation, can easily be added to the model in the form of teacher knowledge. It acts as a good regularizer for the learned embeddings and generated samples. This is particularly useful for natural image datasets, for which off-the-shelf pretrained models are abundant.

\bibliographystyle{splncs04}
\bibliography{egbib}

\section{Appendix}

\subsection{Finetuning the DTN encoder}
\label{sec:dtnfine}

As mentionned in Section \ref{sec:dtn}, the main drawback of DTN is that it keeps a fixed pretrained encoder, which cannot bridge the visual appearance gap between domains.
Following this observation, we perform another experiment in which we finetune the FaceNet encoder relatively to the semantic consistency loss, additionally to the decoder parameters. 

\begin{figure}[!tbh]
\begin{center}
  \begin{minipage}[t]{0.66\textwidth}
    \centering \includegraphics[width=0.485\textwidth, trim={0 0cm 4.5cm 0}, clip]{Images/dtn/dtn_fixed_newtemplate_framed.png}~~
     \includegraphics[width=0.485\textwidth, trim={0 4.87cm 4.7cm 0}, clip]{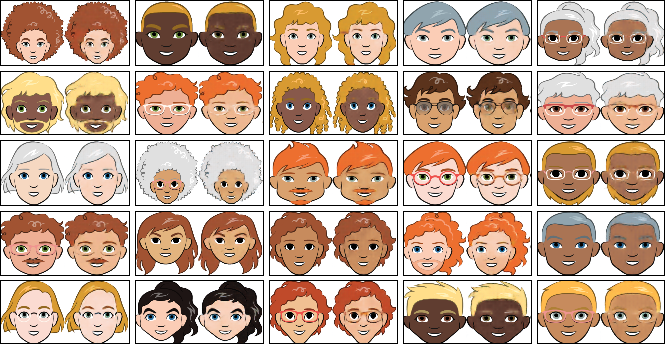}

    \centering \small (a) Random generated samples (left) and reconstructions (right) with fixed FaceNet embedding
  \end{minipage}~~
  \begin{minipage}[t]{0.32\textwidth}
    \centering \includegraphics[width=\textwidth, trim={0 4.87cm 4.7cm 0}, clip]{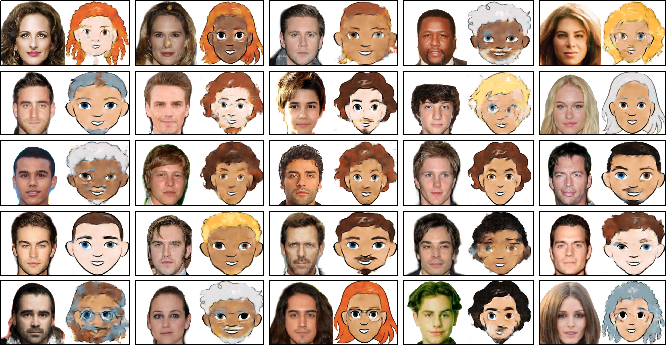}

    \centering \small (b) Random samples with fine-tuned FaceNet encoder
  \end{minipage}
  \end{center}
  \caption{\label{fig:dtnexperiments} Reproducing the Domain Transfer Network performs badly in our experimental setting (a); fine-tuning the encoder yields better results (b) but is unstable for training in practice.}
\end{figure}

While this yields visually better samples (see Figure \ref{fig:dtnexperiments}(b)), it also raises the classical domain adaptation issue of guaranteeing that the initial FaceNet embedding knowledge is preserved when retraining the embedding.
In comparison, \xgan{} exploits a teacher network that can be used to distill prior domain knowledge throughout training, when available.
Secondly, this finetuned DTN is prone to mode collapse. In fact, the encoder is now only trained relatively to the semantic consistency loss which can be easily minizimed by mapping each domain to the same point in the embedding space, leading to the same cartoon being generated for all of them.
In \xgan{}, the source embeddings are regularized by the reconstruction loss on the source domain.
This allows us to learn a joint domain embedding from scratch in a proper domain adaptation framework.

\subsection{Extensive qualitative evaluation}
\label{sec:qual}

\begin{figure}[!htb]
\begin{minipage}[c]{0.24\textwidth}
\begin{center}
\includegraphics[width=\textwidth]{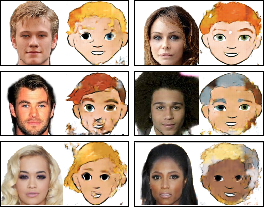}

(a) \textit{Baseline:} DTN
\end{center}
\end{minipage}~
\begin{minipage}[c]{0.24\textwidth}
\begin{center}
\includegraphics[width=\textwidth]{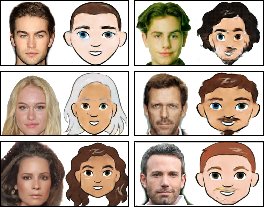}

(b) Finetuned DTN
\end{center}
\end{minipage}
~
\begin{minipage}[c]{0.24\textwidth}
\begin{center}
\includegraphics[width=\textwidth]{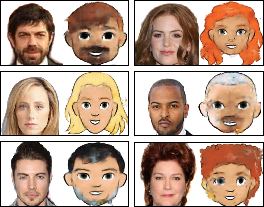}

(c) XGAN ($\mathcal{L}_{r}$ only)
\end{center}
\end{minipage}~
\begin{minipage}[c]{0.24\textwidth}
\begin{center}
\includegraphics[width=\textwidth]{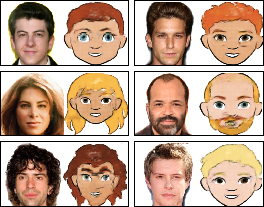}

(d) XGAN
\end{center}
\end{minipage}

\caption{\label{fig:comp} Cherry-picked samples for the DTN baseline and three improved models we consider for the semantic style transfer task}
\end{figure}

As mentioned in the main text, the DTN baseline fails to capture a meaningful shared embedding for the two input domains. Instead, we consider and experiment with three different models to tackle the semantic style transfer problem. Selected samples are reported in Figure \ref{fig:comp}:
\begin{itemize}
\item \textbf{Finetuned DTN}, as introduced previously. In practice, this model yields satisfactory samples but is very sensitive to hyperparameter choice and often collapses to one model.
\item \textbf{XGAN with $\mathcal{L}_{rec}$ and $\mathcal{L}_{dann}$ active only} corresponds to a simple domain-adaptation setting: the proposed \xgan{} model where only the  reconstruction loss $\mathcal{L}_{rec}$ and the domain-adversarial loss $\mathcal{L}_{dann}$ are active. We observe that semantics are globally well preserved across domains although the model still makes some basic mistakes (\eg gender misclassifications) and the samples quality is poor.
\item \textbf{XGAN}, the full proposed model, yields the best visual samples out of the models we experiment on. In the rest of this section, we report a detailed study on its different components and possible failure modes. 
\end{itemize}

In Figure \ref{fig:qual} we also report a more extensive random selection of samples produced by \xgan{}. Note that we only used a discriminator for the source to target path (\ie $\lgants$ is inactive); in fact the GAN objective tends to make training more unstable so we only use one for the transformation we care most about for this specific application, \ie faces to cartoons.
Other than the GAN objective, the model appears to be robust to the choice of hyperparameters. 

Overall, the cartoon samples are visually very close to the original dataset and main identity characteristics such as face shape, hair style, skin tone, etc., are well preserved between the two domains.
The main failure mode appears to be mismatched hair color: in particular, bright red hair appear very often in generated samples which is likely due to its abundance in the training cartoon dataset. 
In fact, when looking at the target to source generated samples, we observe that this color shade often gets mapped to dark brown hair in the real face domain. 
One could expect the teacher network to regularize the hair color mapping, however FaceNet was originally trained for face identification, hence is most likely more sensitive to structural characteristics such as face shape.
More generally, most mistakes are due to the shift in \textit{content} distribution rather than \textit{style} distribution between the two domains. Other examples include bald faces being mapped to cartoons with light hair (most likely due to the lack of bald cartoon faces and the model mistaking the white background for hair color). Also, eyeglasses on cartoon faces disappear when mapped to the real face domain (only very few faces in the source dataset wear glasses).

\subsection{Failure mode when training with $\mathcal{L}_{rec}$ and $\ldann$}
\label{sec:dannfail}

\begin{figure}[!htb]
\begin{center}
   \begin{minipage}[c]{0.45\textwidth}
     \centering \includegraphics[trim={0 4.85cm 9.35cm 0}, clip, width=\textwidth]{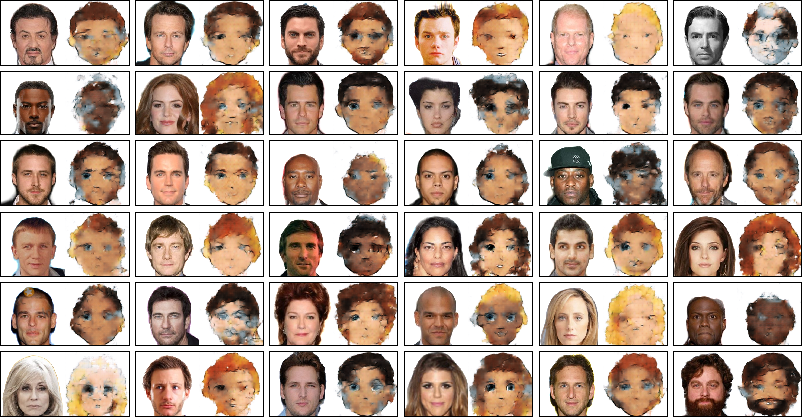}

     \centering \small source to target
   \end{minipage}~~~
   \begin{minipage}[c]{0.45\textwidth}
     \centering \includegraphics[trim={0 4.85cm 9.35cm 0}, clip, width=\textwidth]{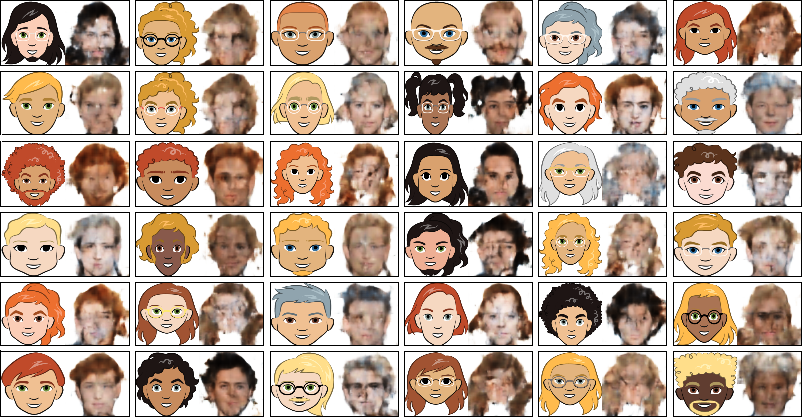}

     \centering \small target to source
   \end{minipage}
   \end{center}
  \caption{\label{fig:dannfail} Random test samples for both cross-domain translations in the failure mode for the $\mathcal{L}_{rec} + \ldann$ only \xgan{} setting}
\end{figure}

In Figure \ref{fig:dannfail} we report examples of failure cases when $\omega_{dann}$ is too high in the setting with the reconstruction and domain-adversarial loss only: the domain-adversarial classifier $c_{dann}$ reaches perfect accuracy and cross-domain translation fails.

\subsection{GAN loss ablation experiment}
\label{app:gan}
\begin{figure}[!tbh]
\begin{center}
  \begin{minipage}[c]{0.47\textwidth}
    \centering \includegraphics[trim={0 4.85cm 9.35cm 0}, clip, width=\textwidth]{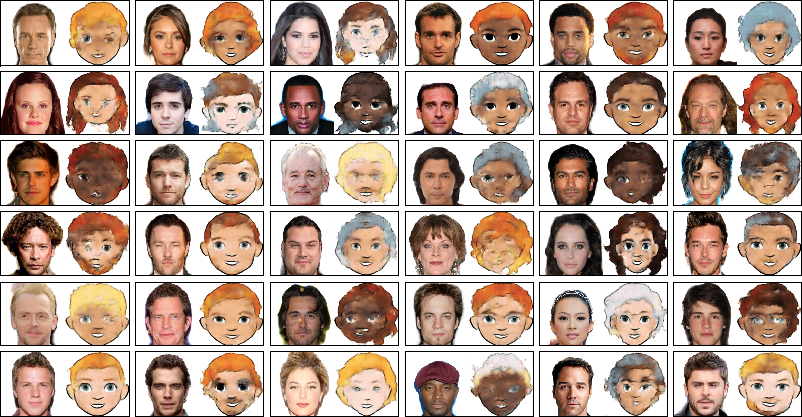}

    \centering \small (a) Without total variation loss
  \end{minipage}~~~
  \begin{minipage}[c]{0.47\textwidth}
    \centering \includegraphics[trim={0 4.85cm 9.35cm 0}, clip, width=\textwidth]{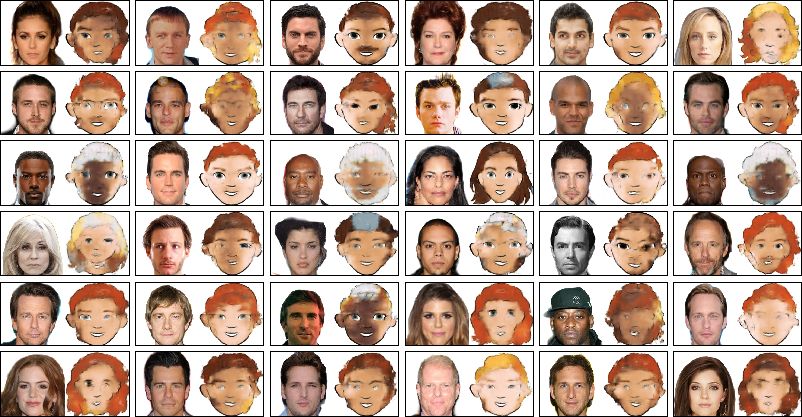}

    \centering \small (b) With total variation loss
  \end{minipage}
  \end{center}
  \caption{\label{fig:nogan} Test samples for \xgan{} when the GAN loss $\mathcal{L}_{ga}$ is inactive}
\end{figure}

As mentioned previously, we only use a GAN loss term for the source $\rightarrow$ target translation, to ease training. This prompts the face-to-cartoon path to generate more realistic samples.
As expected, when the GAN loss is inactive, the generated samples are
noisy and unrealistic (see Figure \ref{fig:nogan}(a)).
For comparison, tackling the low quality problem with simpler regularization techniques such as using total variation smoothness loss leads to more uniform samples but
significantly worsen their blurriness on the long term (see
Figure \ref{fig:nogan}(b)).
This shows the importance of the GAN objective for image generation applications, even though it makes the training process more complex.

\end{document}